\documentclass[a4paper,conference]{IEEEtran}

\usepackage{ifpdf}

\ifCLASSINFOpdf
   \usepackage[pdftex]{graphicx}
   \graphicspath{{../pdf/}{../jpeg/}}
   \DeclareGraphicsExtensions{.pdf,.jpeg,.png}
\else
\fi

\usepackage[hyphens]{url}

\usepackage{siunitx}
\usepackage[super]{nth}
\usepackage{subcaption}
\usepackage{booktabs}
\usepackage[capitalise]{cleveref}
\usepackage[inline]{enumitem}
\usepackage{tabularx}
\usepackage{microtype}
\usepackage[inline]{enumitem}
\setlist*[enumerate]{label=(\arabic*)}

\usepackage{xspace}
\makeatletter
\DeclareRobustCommand\onedot{\futurelet\@let@token\@onedot}
\newcommand{\@onedot}{\ifx\@let@token.\else.\null\fi\xspace}
\makeatother

\newcommand{\etal}{\emph{et~al\onedot}}
\newcommand{\ie}{i.\,e.,\xspace}
\newcommand{\eg}{e.\,g.,\xspace}

\newcommand{\cf}{cf\onedot}

\begin{document}
\title{ODOR: The ICPR2022 \emph{OD}europa Challenge on Olfactory \emph{O}bject \emph{R}ecognition}

\author{\IEEEauthorblockN{Mathias Zinnen\IEEEauthorrefmark{1},
Prathmesh Madhu\IEEEauthorrefmark{1},
Ronak Kosti\IEEEauthorrefmark{1},
Peter Bell\IEEEauthorrefmark{3},
Andreas Maier\IEEEauthorrefmark{1},
Vincent Christlein\IEEEauthorrefmark{1}
}
\IEEEauthorblockA{\IEEEauthorrefmark{1}
Pattern Recognition Lab, Friedrich-Alexander-Universität Erlangen-Nürnberg, 
Germany}
\IEEEauthorblockA{\IEEEauthorrefmark{2}
German Studies and Arts, Philipps-Universität Marburg, 
Germany}
}

\maketitle

\begin{abstract}
The Odeuropa Challenge on Olfactory Object Recognition aims to foster the development of object detection in the visual arts and to promote an olfactory perspective on digital heritage. Object detection in historical artworks is particularly challenging due to varying styles and artistic periods. Moreover, the task is complicated due to the particularity and historical variance of predefined target objects, which exhibit a large intra-class variance, and the long tail distribution of the dataset labels, with some objects having only very few training examples. These challenges should encourage participants to create innovative approaches using domain adaptation or few-shot learning. We provide a dataset of 2647 artworks annotated with 20\,120 tightly fit bounding boxes that are split into a training and validation set (public). A test set containing 1140 artworks and 15\,480 annotations is kept private for the challenge evaluation.
\end{abstract}

\IEEEpeerreviewmaketitle

\section{Introduction}
Cultural heritage has been blind to the olfactory senses of our nose. Olfaction is a crucial element of human experience but has not received much attention in the context of cultural heritage, yet. The Odeuropa project\footnote{\url{www.odeuropa.eu}} aims to remedy this shortcoming by promoting, preserving, and recreating the olfactory heritage of Europe. It is possible to make traces of past smells accessible by automatic analyzing large corpora of visual and textual data from \nth{16} to \nth{20}-century European history. However, finding smell references in historical artworks is a very challenging task. These references can be implicit in a painting's narrative, the actions of depicted characters, or the depicted spaces. We try to approximate the recognition of complex and implicit smell references by first detecting objects with olfactory relevance, based on which more complex smell references might be recognized. The detection of olfactory objects in historical artworks is challenging in multiple aspects:
\begin{figure}
    \centering
    \includegraphics[width=\columnwidth]{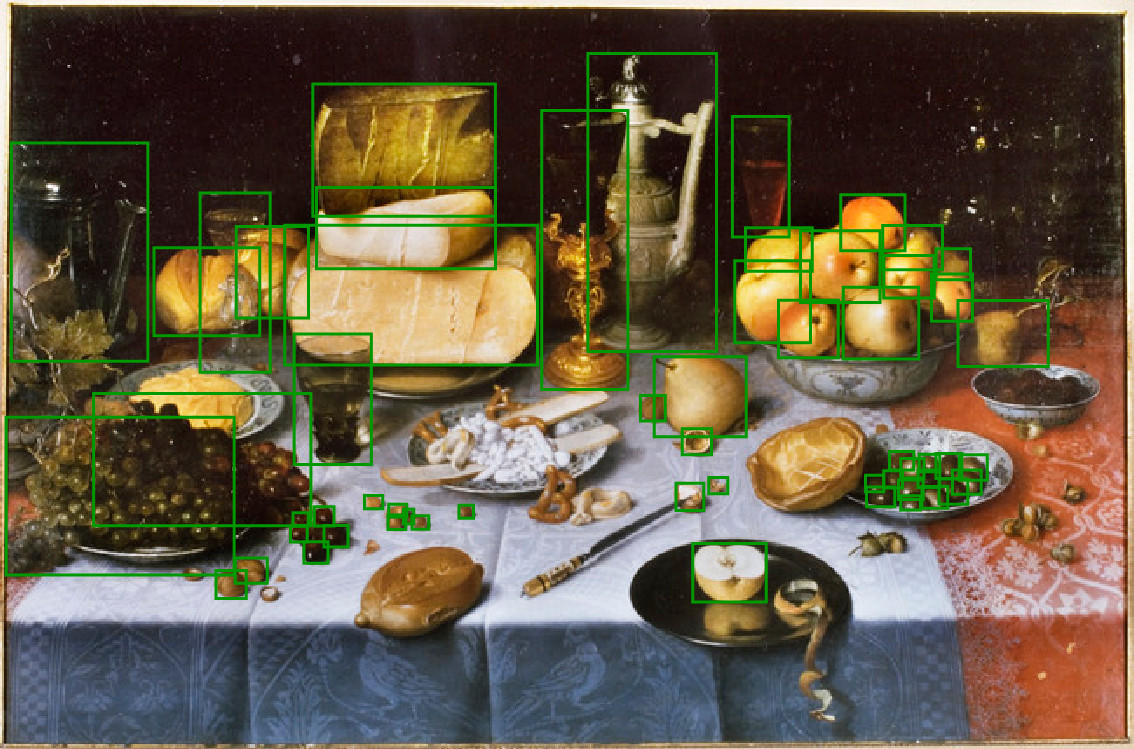}
    \caption{Example image from the challenge dataset exhibiting a large number of small, partially occluded objects. Image credit: \textit{Laid Table with Cheese and Fruit}. 1610. Floris van Dyck. Public Domain, via Wikimedia Commons.}
    \label{fig:example}
\end{figure}
\begin{enumerate}
    \item Object detection in the artistic domain requires algorithms to cope with varying degrees of abstraction and artistic styles, which leads to a considerably higher intra-class variance than photographic depictions.
    \item In contrast to the famous COCO~\cite{lin2014microsoft} and ImageNet~\cite{russakovsky2015imagenet} datasets, where the images usually contain repetitive objects with huge per sample instances, historical artworks usually contain many object instances of diverse sizes, which are often partially occluded (\cf \cref{fig:example}). 
    \item Smell-relevant objects can be particular, leading to a fine-grained classification of target objects. Different types of flowers, for example, might have a different smell although looking very similar.
    \item Since the dataset covers a period over multiple centuries, the appearance of some target objects is subject to historical change. Particularly, man-made objects like cigars or beverages might have changed their look over the years, whereas others like flowers or animals remained mostly invariant.
\end{enumerate}
The category and domain gap between photographic datasets and our target domain poses a challenge that encourages new approaches to increase object detection models' robustness and transfer capability.
In posing the double challenge of overcoming a domain and category gap, we want to foster the development of domain adaptation techniques in object detection and promote a multisensory cultural heritage perspective on computer vision that acknowledges the importance of olfaction.

We allow and encourage the use of different kinds of pre-training on photographic data to enable various domain adaptation methods, \eg transfer learning or style transfer.
Along with our annotated dataset, we provide a hierarchy of object categories, which facilitates the implementation of hierarchical approaches to object detection.

\section{Dataset}

We provide the first dataset of olfactory objects within artworks for the challenge. This section describes the collection, annotation, and a brief description of class distribution. 
\subsection{Image Collection \& Annotation}
As a prerequisite for the assembly of the dataset, we queried multiple digitized museum collections using a list of search terms (\cf \cref{tab:searchterms}) that allegedly led to images with olfactory relevance.
\begin{table}[tb]
       \centering
    \begin{tabular}{lr}
        \toprule
         search term & \# images \\
         \midrule
         Smell\textsuperscript{a}
         & 618
         \\
         Senses\textsuperscript{b}
         & 2217
         \\
         Lazarus\textsuperscript{c}
         & 4215
         \\
         Still Life\textsuperscript{d}
         & 21074
         \\
         Gloves & 901 \\
         Donkey\textsuperscript{e}& 2,483 \\
         Goat & 5,177 \\
         Cheese & 365 \\
         Pomander & 146 \\
         Tobacco & 1,922 \\
         Whale\textsuperscript{f} & 229 \\
         Censer\textsuperscript{g} & 195 \\
         \midrule
         Total & 41,552\textsuperscript{h}\\
         \bottomrule
    \end{tabular}
     \caption{Overview of search terms with the number of images collected for each. 
    \\
    Search term variations: \textsuperscript{a}Geruch, odore, geur, odeur;
    \textsuperscript{b}sens, sensi, Sinne, zintuig;
    \textsuperscript{c}Lazare, Lazarro;
    \textsuperscript{d}natura morta, natura morte, stillleben, stilleben, stilleven;
    \textsuperscript{e}Ezel;
    \textsuperscript{f}Walvis.\\
    \textsuperscript{g}A censer is an incense burner used to burn incense or perfume in solid form.
    }
    \label{tab:searchterms}
\end{table}%
Our image collection strategy is two-fold: In the first step, we defined an initial list of terms that reflect our expectations at the start of the Odeuropa project work, which led to a collection of \num{30134} artworks. 
As our knowledge about contexts in which smell active objects might appear evolves in the annotation process, we iteratively extended the image base with new search terms that have become relevant. 

The objects were annotated manually using cvat\footnote{\url{https://openvinotoolkit.github.io/cvat/}} and Amazon mechanical turk (only flower subcategories). 

We predefined a set of categories that were then iteratively extended resulting in a list of 222 classes to date. 
The high number of object categories, including objects that are very rare and particular, suggests the usage of a hierarchical structure of classes, which has multiple advantages:
\begin{enumerate*}
    \item It makes it easier to find specific object categories, simplifying the annotation process.
    \item Detection systems can resolve to a fallback solution in cases where the exact object category cannot be determined but a broader classification can be made (\eg detecting a flower instead of flower species).
\end{enumerate*}
In contrast to a WordNet-based concept hierarchy like it is applied by Redmon \etal~\cite{redmon2017yolo9000}, we incorporate only two levels of abstraction since a more complex hierarchy remains mostly unused and complicates annotation and detection architectures without adding much extra value. 
From the leave nodes, the complete WordNet hierarchy can, however, still be created. 
The selection of the supercategories is based on pragmatic
considerations such as visual similarity, assumed familiarity with concepts, and simplicity.

Finally, we filtered out supercategories that had less than ten samples for creating the challenge dataset, resulting in a list of 87 categories.

\subsection{Label Distribution}

\Cref{tab:supercats} lists the supercategories that have been used in the annotation scheme and how many subcategories have been defined for each as well as the number of samples in each supercategory.

\begin{table}[h]
    \centering
    \begin{tabular}{lcr}
        \toprule 
         supercategory &  \# subcategories & \# samples\\
         \midrule
         flower & 20 & 8,484\\
         fruit & 28 & 5,196\\
         mammal & 38 & 2,126 \\
         bird & 13 & 1,185\\
         vegetable & 26 & 1,088\\
         smoking equipment & 16 & 958\\
         insect & 17 & 708\\
         beverage & \phantom{0}5 & 553\\
         jewellery & 11 & 433\\
         seafood & 10 & 321\\
         reptile/amphibia & \phantom{0}3 &105\\
         nut & \phantom{0}3 & 78\\
         other & 14 & 1,094 \\
         \bottomrule
    \end{tabular}
    \caption{Supercategories of the annotation scheme.
    The middle column gives the number of subcategories that have been defined for each of the supercategories. 
    The right column reports the number of samples that have been annotated for the supercategory including its subtypes. 
    \emph{Other} subsumes all top-level categories that do not have further subcategories.}
    \label{tab:supercats}
\end{table}

\begin{figure}[t]
    \centering
    \begin{subfigure}{.45\textwidth}
    \includegraphics[width=\textwidth]{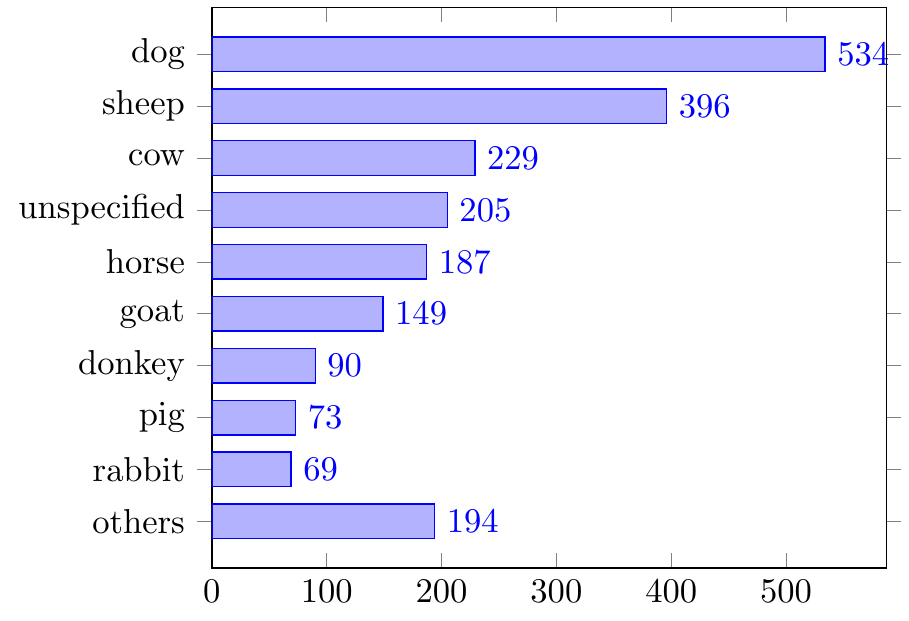}
    \caption{}
    \label{fig:mammals}
    \end{subfigure}
    \quad
    \begin{subfigure}{.45\textwidth}
    \includegraphics[width=\textwidth]{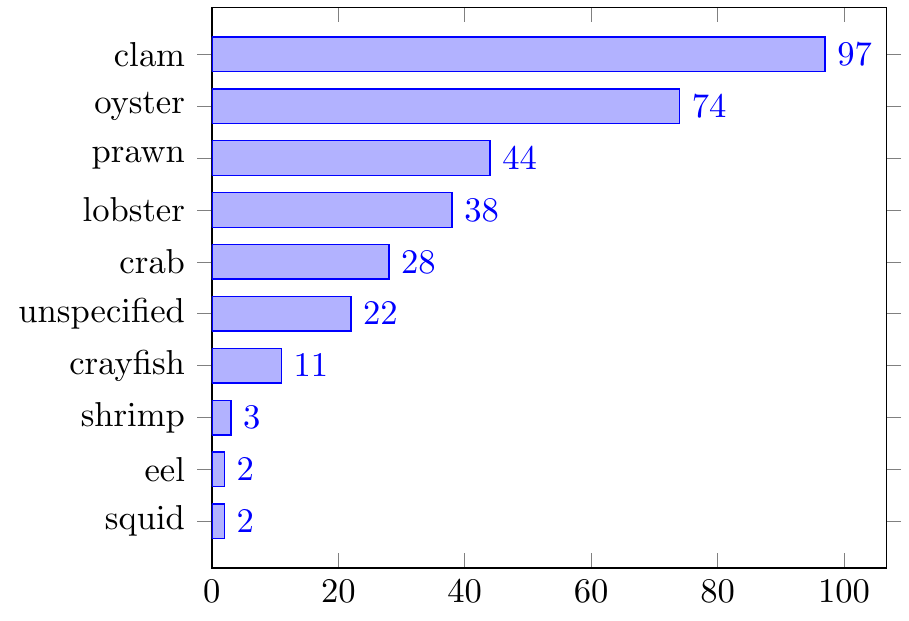}
    \caption{}
    \label{fig:seafood}
    \end{subfigure}
    \caption{Distribution of subcategory annotations of (\subref{fig:mammals}) mammals and (\subref{fig:seafood}) seafood supercategories.}
\end{figure}

\Cref{fig:mammals,fig:seafood} show the exemplary subcategory distributions of the \emph{mammal} and \emph{seafood} categories, respectively.

\subsection{Distribution Format}
For the sake of license compliance, we cannot publish the images directly.
Instead, we provide a CSV file with links pointing to the image sources and a script to conveniently download them.
The annotations are provided in COCO JSON format\footnote{\url{https://cocodataset.org/#format-data}} which defines a bounding box as $[x,y,w,h]$, with $x$ and $y$ denoting the coordinates of the upper left corner of a box, and $w$, $h$ the box width and height, respectively.
Additionally, each bounding box is assigned to one of the predefined categories via the \emph{category\_id} attribute.
Apart from the publication via codalab, the challenge training set has also been published on zenodo~\cite{aidv1} including additional metadata.

\section{Challenge Overview}

\subsection{Challenge Protocol and Duration}
The aim of the ODOR challenge is to locate and classify a diverse range of odor-active objects on historical artworks. The participants are provided with a training set of artwork images along with the bounding box annotations of the target objects. Additionally, they are also provided with a validation set of images without annotations. These images can be used for the algorithm development or model training. The competition started with a preliminary warm-up phase, where the participants were provided with the training data and a starter kit enabling them to perform exploratory data analysis and build initial prototypes and setup their code. The challenge was conducted in two main phases: \begin{enumerate*} \item a development phase \item and a final phase. \end{enumerate*}For both \emph{development} and \emph{final} phase, submissions were expected as a zip file containing the predictions as a COCO-\emph{JSON} format.

\noindent\subsubsection{Development phase.} For the development phase, the bounding box annotations for the validation set were not provided to the participants. During this phase, participants were allowed to upload their predictions on the validation set. The validation set bounding boxes were used to evaluate each participant's submission and provide feedback as per the COCO evaluation metric. Each participant was allowed to upload one submission per day.

\noindent\subsubsection{Final phase.} During the final phase, the validation annotations and the test set (without annotations) were provided to the teams to further fine-tune their models and present robust and generic algorithms on the test set. Similar to the development phase, they were required to submit their results on the test set. For this phase, for each participant, a total of six submissions was allowed.

\subsection{Evaluation Metrics}\label{subsec:evalmetrics}
We use \emph{COCO metric} as the evaluation metric which determines the participants ranking in the final ranking.
To understand any object detection metric, we need to understand Intersection over Union~(IoU).
IoU decides if a predicted bounding box is correct with respect to the ground truth object bounding box or not. 
IoU is defined as the ratio of intersection and union between the predicted and actual bounding box. 
A prediction is considered to be correct (True Positive) if IoU is greater than a predefined threshold value, and False Positive otherwise. 
For COCO evaluation, the predefined IoU thresholds range from 0.5 to 0.95 with a step size of 0.05.
We evaluate \emph{COCO metric} by calculating the mean average precision (mAP) averaged over all classes, averaged over all threshold values (IoU 0.5:0.05:0.95).
Since our dataset contains many small objects that are particularly difficult to detect, we additionally also report the mAP for small, medium, and large objects separately.

\subsection{Participation}
A total of \num{36} teams registered for the challenge, out of which \num{6} teams submitted during the development phase, 
and \num{4} teams submitted their predictions for the final phase. %
Although we are happy with the contribution of the existing participants, we initially expected more submissions.
One reason might be the challenging nature of the dataset which might discourage some scholars.
By skimming through the available codalab challenges, some scholars might have also misinterpreted the challenge name which, in its abbreviated form, does not explicitly link to object detection. 
We plan to create a follow-up where we consider these findings and attract more participants.

\begin{figure*}[t]
\centering
    \begin{subfigure}[t]{0.19\textwidth}
        \centering
        \includegraphics[width=\textwidth, height=8em]{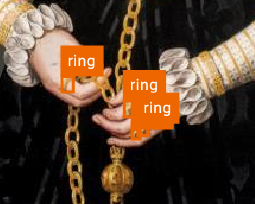}
    \end{subfigure}\hfill
    \begin{subfigure}[t]{0.19\textwidth}
        \centering
        \includegraphics[width=\textwidth, height=8em]{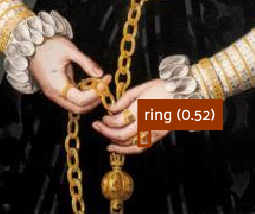}
    \end{subfigure}\hfill
     \begin{subfigure}[t]{0.19\textwidth}
        \centering
        \includegraphics[width=\textwidth, height=8em]{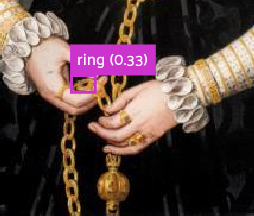}
    \end{subfigure}
    \begin{subfigure}[t]{0.19\textwidth}
        \centering
        \includegraphics[width=\textwidth, height=8em]{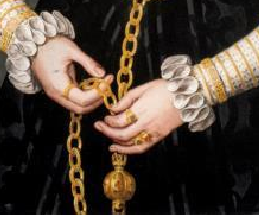}
    \end{subfigure}
    \begin{subfigure}[t]{0.19\textwidth}
        \centering
        \includegraphics[width=\textwidth, height=8em]{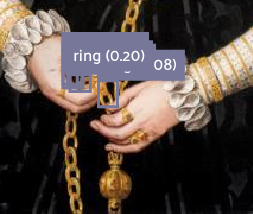}
    \end{subfigure}
    
    \par\medskip
    
    \begin{subfigure}[t]{0.19\textwidth}
        \centering
        \includegraphics[width=\textwidth,height=.1\textheight]{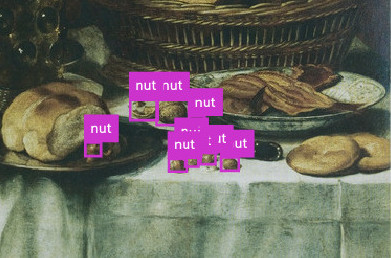}
        \caption{Ground Truth}
    \end{subfigure}\hfill
    \begin{subfigure}[t]{0.19\textwidth}
        \centering
        \includegraphics[width=\textwidth,height=.1\textheight]{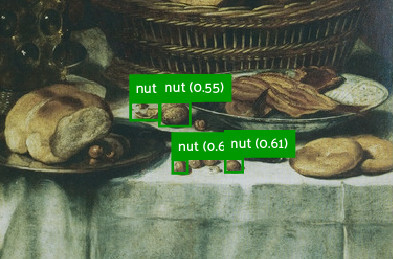}
        \caption{Thousandwords}
    \end{subfigure}\hfill
     \begin{subfigure}[t]{0.19\textwidth}
        \centering
        \includegraphics[width=\textwidth,height=.1\textheight]{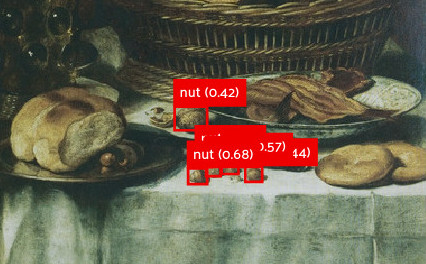}
        \caption{None}
    \end{subfigure}
    \begin{subfigure}[t]{0.19\textwidth}
        \centering
        \includegraphics[width=\textwidth, height=0.1\textheight]{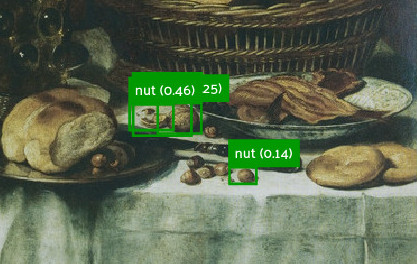}
        \caption{DeadlyDL}
    \end{subfigure}
    \begin{subfigure}[t]{0.19\textwidth}
        \centering
        \includegraphics[width=\textwidth, height=0.1\textheight]{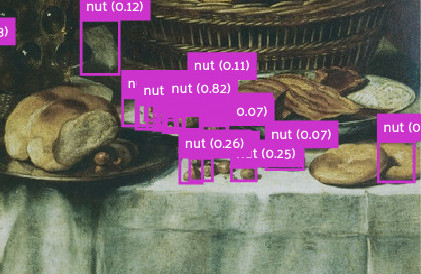}
        \caption{angelvillar96}
    \end{subfigure}
\caption{Qualitative comparison of small-object prediction results of the four finalists. The first row shows predictions for rings in a portrait, whereas the second row shows predictions of partially occluded nuts.\\
Image credits: \textit{(top)}~ \textit{Portrait of an 18-year old woman}. Attributed to Pieter Pourbus. 1574. Oil on panel. RKD – Netherlands Institute for Art History, RKDimages (280945).
\textit{(bottom)}~ Detail from \textit{Stilleven met een mand met kazen}. Pieter Claesz. 1645 -- 1661. Oil on panel. RKD – Netherlands Institute for Art History, RKDimages (108716). } 
\label{fig:qualeval}
\end{figure*}

Team \textit{Thousandwords} consists of Ten Long (University of Amsterdam), Sadaf Gulshad (University of Amstedam), Stuart James (Istituto Italiano di Tecnologia), Noa Garcia (Osaka University), and Nanne von Noord (University of Amsterdam). 
They proposed the use of a strong object detector network called PPYOLO-E~\cite{long2020pp} with a CSP-Resnet~\cite{wang2019cspnet} backbone. 
The final results were obtained by training the network for 150 epochs using a batch size of 10, base learning rate~(LR) of 2.5e-3.
They used stochastic gradient descent with momentum as optimizer for the final model. 
The final model training used a LR scheduler for 5 epochs of \emph{LinearWarmup} and maximum 360 epochs of \emph{CosineDecay}. 
For augmentations, they used \textit{BatchRandomResize} with target random sizes of [320, 352, 384, 416, 448, 480, 512, 544, 576, 608, 640, 672, 704, 736, 768] and random interpolation. 
They normalized the images with a mean of [0.485, 0.456, 0.406] and standard deviation of [0.229, 0.224, 0.225]. 
They experimented with various training schemes like using grayscale images as augmentation, excluding small bounding boxes for robust learning and style transfer as augmentation for domain adaptation. Interestingly however, they report that none of these techniques work better than using a strong object detection model.

Team \textit{None}\footnote{The participants preferred not to be mentioned in the paper.} proposed the use of a YoloV5~\cite{yolov5} model pre-trained on COCO.
They fine-tuned the model using the Ultralytics platform\footnote{\url{https://github.com/ultralytics/}} for 50 epochs with a learning rate of 1e-3 and a batch size of 16. 
For training, the team applied mild data augmentation as given by the \textit{aug\_tfms} function of the albumentations~\cite{info11020125} library. %

Team \textit{DeadlyDL} with the single member Badhan Kumar Das (Siemens Healthineers) used a Faster RCNN~\cite{ren2015faster} model for this task. The model was trained for 80 epochs with a learning rate of c. 5e-4 (0.000478), determined by the learning rate finder~\cite{smith2017cyclical}, a batch size of 2 and ADAM optimizer. For preprocessing, the team used padding and data normalization before passing the images to the neural network. 

Team \textit{angelvillar96} (Angel Villar-Corrales, University of Bonn) used a single-shot object detection network called RetinaNet~\cite{lin2017focal} with a Resnet50-FPN~\cite{lin2017feature} backbone pretrained on COCO-2017 dataset. The team used  the Adam optimizer with an initial learning rate of 3e-4 with a decay factor of 10 (3e-5, 3e-6). 
The batch size was set to 32 due to hardware limitations and the network was trained for 50 epochs, with the best performance at \nth{45} epoch. The final model was trained on a machine with an NVIDIA RTX A6000 with 48GB. Training for 50 epochs took about 1.5 hours.

To simplify participation, we provided a simple \textit{baseline method} that was published on GitHub\footnote{\url{https://github.com/Odeuropa/ICPR-ODOR-starting-kits/}}. 
For the baseline, we used an ImageNet pre-trained Faster-RCNN with a Resnet-50 FPN backbone. 
First, we fine-tuned only the head for 10 epochs using a learning rate of 1e-3, followed by 50 epochs of training the whole network with the same learning rate of 1e-3 before using a lower learning rate of 1e-4 for another 50 epochs.
Similar to team None, we used mild data augmentation as provided by the albumentation library and normalized the input using ImageNet-based mean and standard deviation.

\section{Challenge Results}
The submissions are ranked according to the \emph{COCO metric}. 
The winner is team \emph{Thousandwords} with members from the University of Amsterdam, Istituto Italiano di Tecnologia, and the Osaka University. 

Second place goes to team \emph{None}. 
\emph{DeadlyDL} from Siemens Healthineers achieves the 3rd place. 
\emph{Angelvillar96} from the University of Bonn scores the 4th place.

In order to comprehensively evaluate the submissions, we also report the mean average precision~(mAP) for small, medium and large bounding boxes. 
As expected, all submissions were struggling with small boxes. 
We can see that \emph{Thousandwords} achieved the highest mAP for all three types of bounding boxes.
As expected, all submissions were struggling with small boxes. 
Compared with middle-sized boxes, we observe a performance decrease of more than 100\% for the first and second ranked team, and an even higher drop of c. 350\% for the other participants. 
\begin{table}[t]
    \centering
    \begin{tabular}{cccc}
    \toprule
         & COCO mAP(\%) & mAP@.5(\%) & mAP@.75(\%) \\
         \midrule
         baseline & 3.99 & 8.92 & 2.95 \\
         \midrule
        Thousandwords & 11.49 & 18.93 & 12.00 \\ 
        None & 7.52 & 12.16 & 8.29 \\
        DeadlyDL & 4.58 & 10.00 & 3.77 \\
        angelvillar96 & 3.82 & 8.41 & 2.65 \\
        \bottomrule
    \end{tabular}
    \caption{Results on the final test set in terms of COCO mAP, Pascal VOC mAP (mAP@.5), and strict evaluation (mAP@.75).}
    \label{tab:results}
\end{table}

\begin{table}[t]
    \centering
    \begin{tabular}{cccc}
    \toprule
         & mAP-small(\%) & mAP-medium(\%) & mAP-large(\%) \\
         \midrule
         baseline & 1.07 & 3.50 & 10.25 \\
         \midrule
        Thousandwords & 4.19 & 11.71 & 25.24 \\ 
        None & 3.03 & 7.36 & 15.74 \\
        DeadlyDL & 1.00 & 4.50 & 10.43 \\
        angelvillar96 & 0.84 & 3.76 & 9.19 \\
        \bottomrule
    \end{tabular}
    \caption{Evaluation of COCO mAP for different object sizes}
    \label{tab:sizes}
\end{table}

\section{Discussion}
\begin{figure*}[t]
    \begin{subfigure}[t]{0.19\textwidth}
        \centering
        \includegraphics[width=\textwidth,height=.1\textheight]{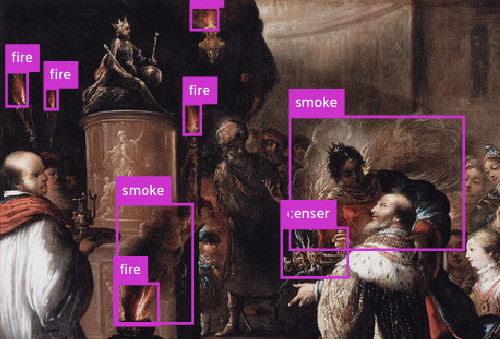}
        \caption{Ground Truth}
    \end{subfigure}\hfill
    \begin{subfigure}[t]{0.19\textwidth}
        \centering
        \includegraphics[width=\textwidth,height=.1\textheight]{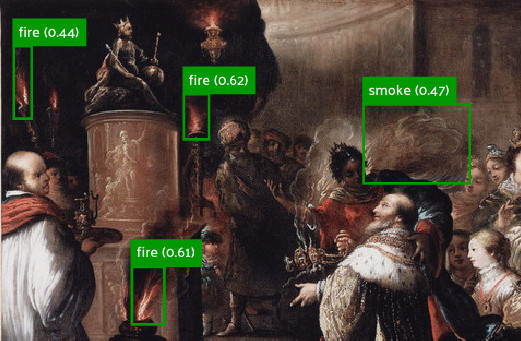}
        \caption{Thousandwords}
    \end{subfigure}\hfill
     \begin{subfigure}[t]{0.19\textwidth}
        \centering
        \includegraphics[width=\textwidth,height=.1\textheight]{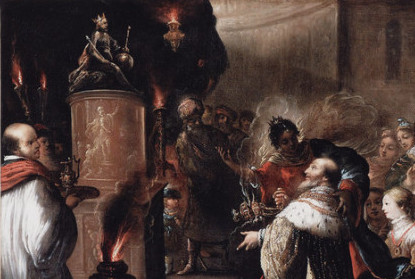}
        \caption{None}
    \end{subfigure}
    \begin{subfigure}[t]{0.19\textwidth}
        \centering
        \includegraphics[width=\textwidth,height=.1\textheight]{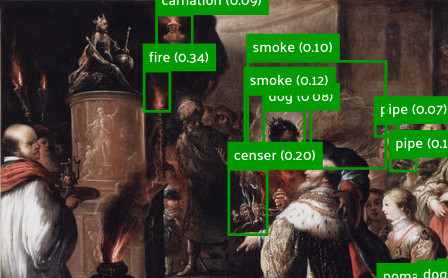}
        \caption{DeadlyDL}
    \end{subfigure}
    \begin{subfigure}[t]{0.19\textwidth}
        \centering
        \includegraphics[width=\textwidth,height=.1\textheight]{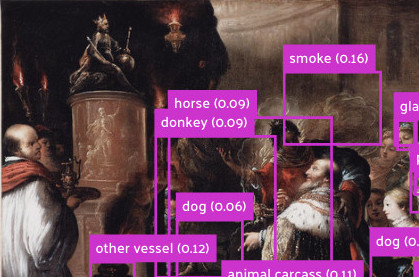}
        \caption{angelvillar96}
    \end{subfigure}
\caption{Qualitative comparison of prediction results for the challenging smoke and fire categories.\\
Image credit: Detail from \textit{Solomon's idolatry (1 Kings 11:7--8)}. Circle of Claude Vignon. 1650--1674. Oil on canvas. RKD – Netherlands Institute for Art History, RKDimages (114441).}
\label{fig:smokefire}
\end{figure*}

The major challenges within this competition were detecting objects, that were less represented in the training data, small objects; periodically changing objects with varying styles, and same class of objects obstructing and overlapping with each other.
\Cref{fig:qualeval} gives examples for some challenging categories. 
The first two rows visualize detections of \emph{small objects}, \ie a portrait with three rings in the first row, and a still-life containing a large number of (partially occluded) nuts in the second row. 
Considering the object size, both nuts and rings are reasonably well detected by team Thousandwords and None.
While the models of the teams DeadlyDL and angelvillar96 seem to largely overestimate the number of instances, the confidence score is below 0.5 for all instances, meaning that the false predictions do not decrease the COCO metric.
However, the large number of overlapping predictions suggest that the usage or modification of non-maximum-suppression might improve the results.
What surprised us was the detection performance for the allegedly challenging categories of smoke and fire.
We expected both categories to be very challenging to detect since, especially in the case of smoke, they lack clear boundaries and their localisation is ambiguous. 
As \cref{tab:firesmoke} shows, our expectation was met for the teams None and angelvillar96 who both achieved a 0.0 precision for these categories.
Surprisingly however, the teams Thousandwords and DeadlyDL achieved precision values considerably higher than their average over all categories.
\Cref{fig:smokefire}, where the Thousandwords and DeadlyDL models both detect instances of smoke with blurry boundaries, emphasizes this finding.

\begin{figure}[t]
\begin{tabularx}{.4\textwidth}{lcc}
       \includegraphics[width=.15\textwidth]{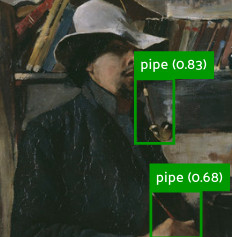} &
       \includegraphics[width=.15\textwidth]{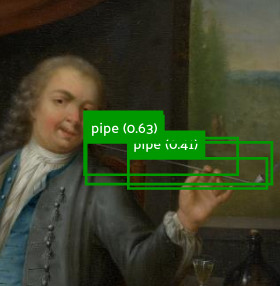} & %
       \includegraphics[width=.15\textwidth]{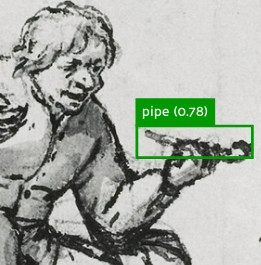}
\end{tabularx}
\caption{Exemplary pipe detections of the winning model over different stylistic representations.\\
Image credits: \textit{(l)} Detail from \textit{Self portrait in the studio}. Jan Toorop. 1883. Oil on panel. RKD – Netherlands Institute for Art History, RKDimages (128870).
\textit{(m)} Detail from \textit{Portrait of a man smoking}. Anonymous. 1800--1850. Oil on panel. RKD – Netherlands Institute for Art History, RKDimages (294941).
\textit{(r)} Detail from \textit{Peasant seated with pipe}. Adriaen van Ostade. 1625--1685. Graphite on paper. RKD – Netherlands Institute for Art History, RKDimages (198724).}
\label{fig:pipes}
\end{figure}
Another positive surprise was the robustness of the participants towards deviations in stylistic representation of the target objects. 
\Cref{fig:pipes} shows detection of the Thousandwords method for three different representations of pipes. 
Although the right image exhibits a completely different artistic style, the pipe detection is still detected successfully.
Furthermore, the different variations of the pipe object exhibited by the leftmost and the middle image seem not to prevent a successful detection.

\begin{table}[t] 
    \centering
    \begin{tabular}{ccc}
			\toprule
         & smoke AP & fire AP  \\
         \midrule
        Thousandwords & 0.44 & 0.33 \\
        None & 0.00 & 0.00 \\
        DeadlyDL & 0.12 & 0.20 \\
        angelvillar96 & 0.00 & 0.00 \\
        \bottomrule
    \end{tabular}
    \caption{Average precision of smoke and fire categories for all finalists. All precision values are reported according to COCO evaluation.}
    \label{tab:firesmoke}
\end{table}

\begin{figure*}[t]
\centering
\begin{subfigure}[t]{0.24\textwidth}
    \centering
    \includegraphics[width=\textwidth,height=.1\textheight]{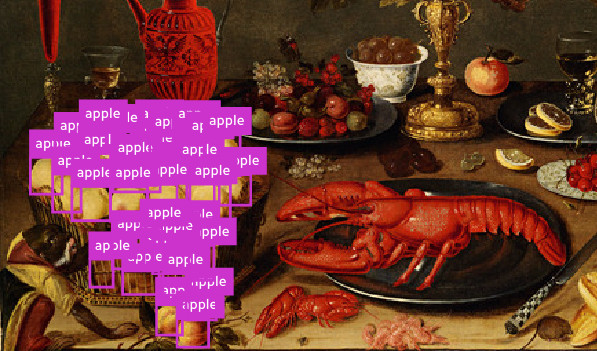}
    \caption{Ground Truth}
\end{subfigure}
\begin{subfigure}[t]{0.24\textwidth}
    \centering
    \includegraphics[width=\textwidth,height=.1\textheight]{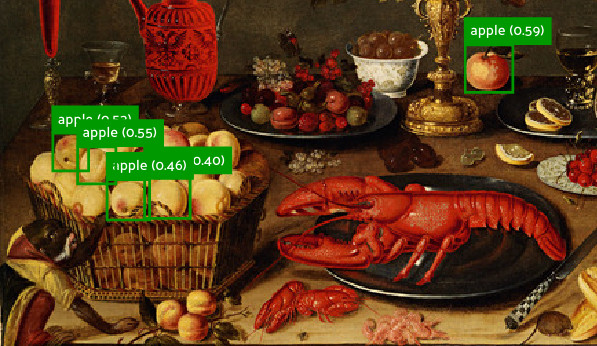}
    \caption{Thousandwords}
\end{subfigure}
\begin{subfigure}[t]{0.24\textwidth}
    \centering
    \includegraphics[width=\textwidth,height=.1\textheight]{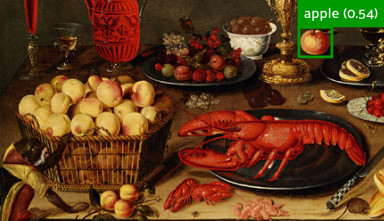}
    \caption{DeadlyDL}
\end{subfigure}
\begin{subfigure}[t]{0.24\textwidth}
    \centering
    \includegraphics[width=\textwidth,height=.1\textheight]{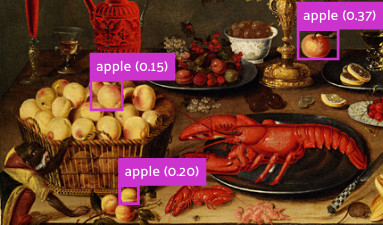}
    \caption{angelvillar96}
\end{subfigure}
\caption{Apple detections on a heap of occluded and overlapping apple instances. Team None did not have any detections.
Image credits: Detail from \textit{Still life with a lobster, glasswork, bread, cheese and parrots}. Artus Claessens. 1615--1644. Oil on canvas. RKD – Netherlands Institute for Art History, RKDimages (16311).}

\label{fig:apples}
\end{figure*}

Challenging as expected was the detection of large numbers of objects partially occluding each other. 
\Cref{fig:apples} shows detections of a heap of apples for three participants.
None of the participant models managed to find the majority of the apples in the heap.
This motivates an evaluation approach similar to the OpenImages~\cite{kuznetsova2020open} evaluation protocol where groups of objects with at least five overlapping instances are counted as successful detections if at least one instance in the bonding box around the group is being detected.
We might adapt this evaluation protocol in a possible future challenge.
Interestingly, we do not observe a confusion between the visually relatively similar categories of apples, peaches, and pears, which is reflected in the confusion matrix between those categories (cf \cref{tab:conf}). %

\begin{table}[t]
    \centering
    \begin{tabular}{lccccc}
    \toprule
         & apple & pear & peach & none & other  \\
         \midrule
        apple & 6 & 0 & 0 & 133 & 0 \\
        pear & 0 & 0 & 0 & 34 & 0 \\
        peach & 0 & 0 & 0 & 11 & 66 \\
        \bottomrule
    \end{tabular}
    \caption{Confusion matrix for detections of apples, pears, and peaches for team Thousandwords}
    \label{tab:conf}
\end{table}

\section{Conclusion}

We held the Odeuropa Challenge on Olfactory Object Recognition to promote object detection in the challenging domain of digital heritage. 
A total of 36 teams participated in the challenge of which 6 submitted to the development phase, and 4 teams submitted to their final predictions.
By raising the attention of digital humanities and computer vision alike, the challenge increased the respective visibility and cooperation.
Particularly in the emerging discipline of olfactory heritage studies, we hope to promote an interdisciplinary approach that considers computational methods. 

We briefly introduced the four final submissions and analyzed their results qualitatively and quantitatively. 
Especially the winning team shows some promising results in terms of small object detection and robustness towards different styles.
To further monitor the progress and enable easy benchmarking of newly developed algorithms, we will reopen the challenge for new submissions.

\section*{Acknowledgment}
For feedback, guidance, professional and moral support we would like to thank Lizzie Marx, Sofia Ehrich, William Tullett, Hang Tran, Inger Leemans, Arno Bosse, Marieke van Erp, the whole Odeuropa Team, and of course all participants.
We gratefully acknowledge the support of NVIDIA Corporation with the donation of the two Quadro RTX 8000 used for this research. The paper has received funding by Odeuropa EU H2020 project under grant agreement No. 101004469.

\bibliographystyle{IEEEtran}
\bibliography{report}

\end{document}